\documentclass[runningheads]{llncs}
\usepackage[T1]{fontenc}
\usepackage{graphicx}
\usepackage{amsmath}
\usepackage{amssymb}
\usepackage{multirow}
\usepackage[pagebackref,breaklinks,colorlinks]{hyperref}

\usepackage{booktabs}
\begin{document}
\title{Masks and Manuscripts: Advancing Medical Pre-training with End-to-End Masking and Narrative Structuring}
%
%
\author{Shreyank N Gowda\inst{1} \and
David A. Clifton\inst{1,2}}
%
\authorrunning{Shreyank N Gowda and David A. Clifton}
%
\institute{Department of Engineering Sciences, University of Oxford, OX3 7DQ Oxford, UK \and
Oxford Suzhou Centre for Advanced Research, University of Oxford, Suzhou 215123, Jiangsu, China}
\maketitle              
\begin{abstract}
Contemporary medical contrastive learning faces challenges from inconsistent semantics and sample pair morphology, leading to dispersed and converging semantic shifts. The variability in text reports, due to multiple authors, complicates semantic consistency. To tackle these issues, we propose a two-step approach. Initially, text reports are converted into a standardized triplet format, laying the groundwork for our novel concept of ``observations'' and ``verdicts''. This approach refines the {Entity, Position, Exist} triplet into binary questions, guiding towards a clear ``verdict''. We also innovate in visual pre-training with a Meijering-based masking, focusing on features representative of medical images' local context. By integrating this with our text conversion method, our model advances cross-modal representation in a multimodal contrastive learning framework, setting new benchmarks in medical image analysis.

\end{abstract}
\section{Introduction}
\label{sec:intro}

Deep learning has seen significant achievements in image classification~\cite{colornet,resnet} and action recognition~\cite{smart,jhuang2013towards,gowda2017human}, driving forward the capabilities of computer vision applications. With the swift advancements in deep learning, computer-aided diagnosis in the medical field has made notable progress through various models. However, these models, predominantly trained on specific anatomical or disease categories, demand costly data annotation and re-training when a new disease emerges~\cite{ji2023unsupervised}, limiting their practical utility. While deep learning has flourished due to large-scale labeled natural image datasets~\cite{imagenet}, the annotation of medical images remains time-consuming and expensive. A common strategy involves pre-training on large datasets like ImageNet~\cite{imagenet} and then fine-tuning on specialized medical datasets~\cite{wen2021rethinking}. While research has shown excellent performance in the natural image~\cite{zsi1,zsi2,zsi3} using this strategy, it often falls short in terms of generalized performance due to the domain differences between natural and medical images.

Medical image analysis stands at the crossroads of technological advancements and clinical applications. The rapid growth of medical imaging modalities and the increasing volume of patient data offer unprecedented opportunities to leverage sophisticated machine learning techniques. Yet, the complexity of this data—rich in variability, granularity, and intricacies—poses unique challenges for the development of effective and reliable models. Adding to the challenge, the associated textual data, which often accompanies these images in the form of clinical notes and radiological reports, can introduce inconsistencies due to varied semantics and writing styles across different authors~\cite{flanders2012radiology}, potentially leading to significant noise in the learning process and compromising model reliability.

Self-supervised learning has emerged as a powerful paradigm for pre-training deep neural networks without relying on human annotations, leveraging inherent structure in the data to learn useful representations in a scalable manner despite their associated costs~\cite{radford2021learning,wattforwhat}. Drawing on these advancements, this paper introduces a novel approach that combines self-supervised learning with the nuanced requirements of medical data processing. We propose a two-step conversion process for textual data into a standardized triplet format to refine representations and an innovative visual pre-training strategy that employs Meijering-based masking for improved feature extraction~\cite{meijering2004neurite}.

Furthermore, vision-language pre-training (VLP) models like CLIP~\cite{radford2021learning} have shown success in learning joint multimodal representations but struggle with the unique challenges presented by medical images and unstructured reports~\cite{medklip}. We address these challenges by employing a filtering-based approach for preprocessing medical images, enhancing the performance of deep learning models through classical techniques like Meijering's vesselness filter~\cite{meijering2004neurite,sato} and adapting these for novel masking strategies to facilitate cleaner reconstructions and improved model performance.

Incorporating external knowledge into medical models has shown promise in mimicking medical decision-making processes and enhancing model training with domain-specific clinical knowledge~\cite{wang2018tienet}. Building on the concept of MedKLIP~\cite{medklip}, we extend the extraction of key data from radiology reports into triplets and encode it with domain knowledge to improve masked generative pre-training~\cite{medklip}.

This paper aims to bridge the gap between the raw, multifaceted medical data and the refined, consistent representations needed for effective machine learning applications, pushing the boundaries of medical pre-training by tailoring advanced techniques specifically for the medical domain. Through an amalgamation of self-supervised learning, VLP models, preprocessing techniques, and the incorporation of medical knowledge, we propose a comprehensive framework for enhancing the performance and reliability of machine learning models in medical image analysis and report generation.

\section{Methodology}

\begin{figure*}
    \centering
    \includegraphics[width=0.95\textwidth]{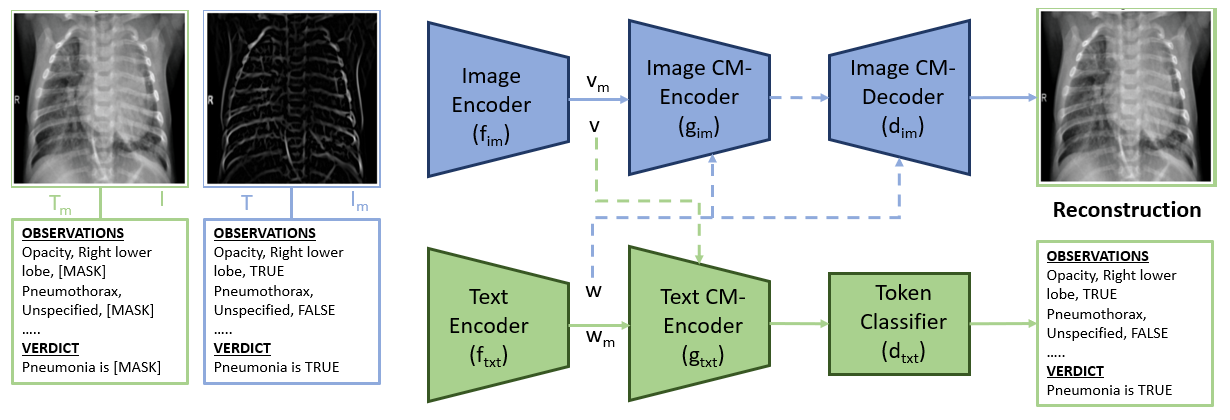}
    \caption{An overview of the architecture for integrated modeling of masked medical visual and linguistic data. The blue and green pathways represent the flow of information for the reconstruction of images and text, respectively. The dashed lines show the intermodal contribution of exposed signals for the creation of attention.}
    \label{fig:overview}
\end{figure*}

In Figure~\ref{fig:overview}, we outline our approach, adopting MaskVLM's~\cite{maskvlm} architecture. Our innovations primarily lie in visual and textual data pre-processing, rather than the architecture itself, which serves as a foundation for our training methodology. We employ transformer-based encoders~\cite{vit,transformer} for concurrent image and text analysis. The image encoder extracts features as vector sets, including a class token, while the text encoder processes text with a start token. Through self-attention and cross-modality encoders featuring cross-attention, we enhance inter-modal integration, culminating in accurate image descriptions. Further details and the rationale for our methods are elaborated in subsequent sections.

\subsection{Masked Image Modeling}

\begin{figure}
    \centering
    \includegraphics[width=0.8\columnwidth]{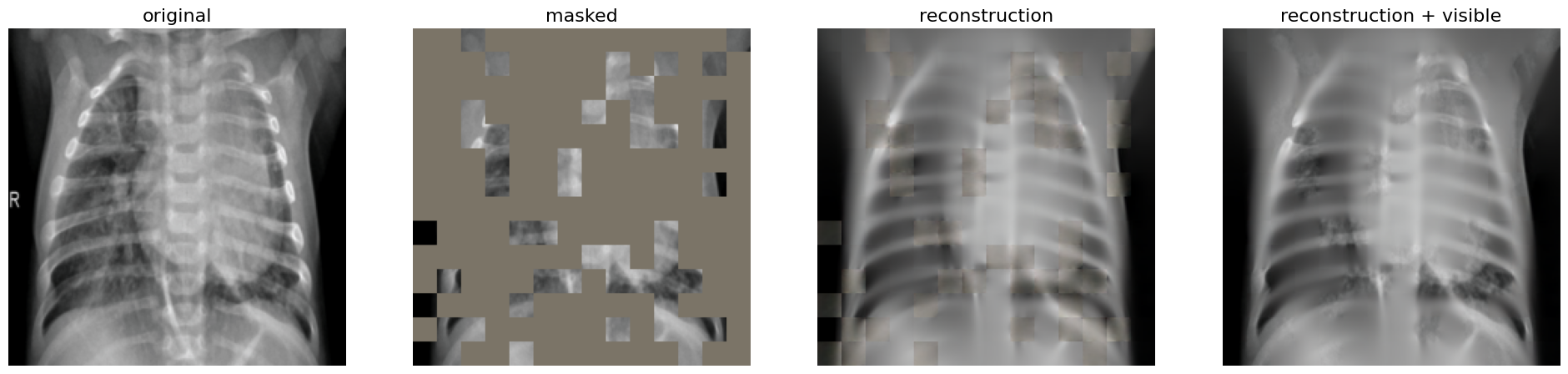}
    \caption{Comparing masking strategies: (a) Masking random patches for reconstruction often results in blurry outputs lacking in detail. (b) Filtering the image before reconstruction preserves fine-grained details, leading to higher resolution outcomes.}
    \label{fig:masking}
\end{figure}

Masked Image Modeling (MIM)~\cite{masked} revolutionizes self-supervised learning by enabling models to infer visual contexts without extensive labeled data, thereby enhancing efficiency and generalization. However, for fine-grained medical data, random masking leads to unclear reconstructions. Despite being superior to random initialization~\cite{mimi}, improvements are needed due to the reliance on local features in medical images. Utilizing ridge filters, like the Meijering filter, converts X-ray images into intermediate forms for reconstruction, addressing the fine-grained nature of medical data~\cite{meijering2004neurite,sato}. This approach, supported by qualitative analysis in Figure~\ref{fig:masking} and further in Table~\ref{tab:manuscript} and Table~\ref{tab:masking_strategy}, empirically outperforms random masking as evidenced by our ablation study.

\subsubsection{Meijering Filter Overview}
The Meijering filter enhances tubular structures in images via the Hessian matrix's eigenvalues~\cite{meijering2004neurite}. For an image \( I(x,y) \), the Hessian matrix \(\mathbf{H}\) and its eigenvalues \( \lambda_1, \lambda_2 \) guide the filter response \( R \), which is zero for \( \lambda_2 > 0 \) and \( \sqrt{\lambda_1^2 + \lambda_2^2} \) otherwise. By adjusting the scale \( \sigma \), the filter adapts to features of various sizes, optimizing the response \( R_{\sigma} \) across scales to highlight relevant structures, thus aiding in medical image analysis.

\subsection{Manuscript Generation}

\begin{figure}
    \centering
    \includegraphics[width=\columnwidth]{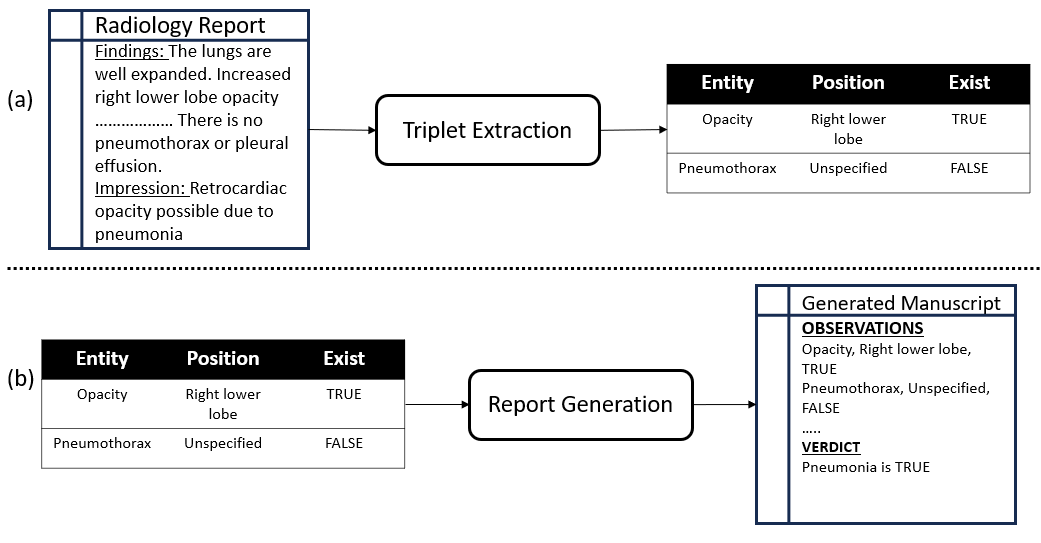}
    \caption{Our report generation process starts with the triplet extraction method from MedKLIP, as outlined in part (a). Instead of adopting the Knowledge-enhanced Triplet Encoding, we transform these into a textual report, embedding it sentence by sentence. This approach facilitates masked pre-training by providing binary labels for each observation and the verdict.}
    \label{fig:textgen}
\end{figure}

Radiology reports in publicly available datasets vary due to individual writing styles, affecting the consistency of sentence embeddings and their use in contrastive learning. We adopt triplet extraction from MedKLIP~\cite{medklip} for uniformity, converting extracted triplets into new reports. This approach ensures semantic consistency, facilitating the extraction of positive and negative pairs, and allows for binary answers in observations and verdicts, useful for masked pre-training. Figure~\ref{fig:textgen} provides an overview of our methodology.

\subsubsection{Report Pre-processing}

Following MedKLIP~\cite{medklip}, we apply RadGraph~\cite{radgraph}, a medical Named Entity Recognition (NER) tool, to streamline radiology reports into triplets. The NER identifies and classifies medical terms as "entity" (clinical findings) or "position" (body location), and assigns an "exist" status to indicate the presence, absence, or uncertainty of clinical observations. This process, generating triplets like {entity, position, exist}, simplifies sentence structures in reports while retaining crucial information, leveraging the Triplet Extraction method in medical Visual Language Processing (VLP).

\subsection{Conditional Reconstruction}

Our method aims to reconstruct an original signal in one modality (image or text) from its masked version using unmasked information from the other modality. We follow notations used by MaskVLM~\cite{maskvlm} as we build on them. Specifically, for a medical image $I$ and its masked form $I_{m}$, with accompanying unmasked text $T$, the goal is to retrieve $I$ utilizing ($I_{m}$, $T$). In reverse, with an unmasked image $I$ and masked text $T_{m}$, the objective is to recover $T$ with $(I, T_{m})$. We employ image and text encoders $f_{im}$ and $f_{txt}$ for feature extraction. The cross-modality encoders $g_{im}$ and $g_{txt}$, leveraging context from unmasked modalities, reconstruct the signals through decoders $d_{im}$ and $d_{txt}$ for images and texts, respectively. This reconstruction process is optimized by a multimodal joint modeling loss $\mathcal{L}_{MVLM}$, promoting focus on unmasked modality features for accurate reconstruction:
\begin{align}
\mathcal{L}_{\text{MVLM}} = & \mathbb{E}_{(I,T)\sim \mathcal{D}} \left[ \mathcal{H} \left( y^M_{T}, \phi_{\text{txt}}^M(I,T|m) \right) + \left\| \frac{1}{\Omega(I^M)} \left( I^M - \phi^M_{\text{im}}(I_m,T') \right) \right\|_1 \right]
\end{align}
where $\phi_{\text{txt}}$ and $\phi_{\text{im}}$ represent the decoding processes for text and image modalities, respectively, focusing on masked regions. This process enables learning from cross-modal interactions to enhance reconstruction accuracy.

\subsection{Multi-Modal Alignment}
To enhance masked signal reconstruction, we incorporate cross-modal alignment through two additional objectives. The first, image-text contrastive (ITC) learning~\cite{radford2021learning}, projects image and text encoder outputs into a shared space, using separate fully-connected layers. This process is formulated as minimizing the contrastive loss $\mathcal{L}_{ITC}$, encouraging agreement between matched image-text pairs in embedding space. However, variability in radiology report styles may affect the identification of positive and negative pairs, as observed in our ablation study.

The ITC loss for the k-th image-text pair, with batch size $N$ and temperature $\tau$, is given by:
\begin{align}
\mathcal{L}_{ITC} = -\frac{1}{N} \sum_{k=1}^{N} \left[ \log \frac{\exp(z^k_{\text{im}} \cdot z^k_{\text{txt}} / \tau)}{\sum_{n=1}^{N} \exp(z^n_{\text{im}} \cdot z^k_{\text{txt}} / \tau)} + \log \frac{\exp(z^k_{\text{im}} \cdot z^k_{\text{txt}} / \tau)}{\sum_{n=1}^{N} \exp(z^k_{\text{im}} \cdot z^n_{\text{txt}} / \tau)} \right],
\end{align}

The second objective, image-text matching (ITM) classification~\cite{itm}, predicts the alignment of image-text pairs by element-wise multiplication of cross-modality encoder outputs, followed by classification. The ITM loss $\mathcal{L}_{ITM}$ ensures supervised alignment across modalities:

\begin{equation}
    \mathcal{L}_{ITM} = \mathbb{E}_{(I,T) \sim D} \left[ \mathcal{H}(y_{ITM}, g^{itm}_{\text{cross}}(z^{cross}_{\text{im}}, z^{cross}_{\text{txt}})) \right].
\end{equation}

The total pre-training loss $\mathcal{L}$ combines all losses:
\begin{equation}
    \mathcal{L} = \mathcal{L}_{MVML} + \mathcal{L}_{ITC} + \mathcal{L}_{ITM}.
\end{equation}

Model fine-tuning is performed for each downstream task over a few epochs, alongside zero-shot evaluations following prior work~\cite{prior,medklip}.

\section{Experimental Analysis}

\subsection{Datasets}

We use datasets that have been used in recent approaches~\cite{gloria,prior,medklip} namely: MIMIC-CXR v2~\cite{johnson2019mimic}, RSNA Pneumonia Detection~\cite{shih2019augmenting}, SIIM-ACR Pneumothorax~\cite{pneumo}, NIH Chest X-Ray Dataset~\cite{nih}, CheXpert~\cite{chexpert}, COVIDx CXR~\cite{covid} and Edema Severity~\cite{edema}. We follow the evaluation protocol of MedKlip~\cite{medklip}. 

\subsection{Implementation Details}

The model utilizes a ViT-B pre-trained on ImageNet for image processing and ClinicalBERT for text, resizing images to 224x224 and setting a common latent dimension of 768. Training is conducted on 4 NVIDIA Tesla V100 GPUs with a batch size of 128, using AdamW optimizer with a 0.05 weight decay. Unlike fixed masking ratios, the model adapts the mask ratio based on the input, using a 15\% ratio for text that increases to 30\% when paired with images. The model is pre-trained for 100 epochs, considerably less than the 800 epochs in similar studies, and fine-tuned for 10 epochs with a specific learning rate strategy, including a warm-up to 3e-4 followed by a cosine scheduler for the pre-training phase, and fixed learning rates of 1e-5 during fine-tuning. Extending training to 300, 500, or 800 epochs did not improve accuracy.

\subsection{Classification}

\subsubsection{Semi and Fully Supervised}

We conduct semi and fully supervised classification on RSNA Pneumonia, SIIM-ACR, and CheXpert datasets, varying labeled data from 1\% to 100\%. Results, including mean and standard deviation over 5 runs using PRIOR~\cite{prior}, are in Table~\ref{tab:supervised_classification}. M\&M surpasses SOTA by up to 2.64\%.

\begin{table*}
    \centering
\resizebox{0.75\textwidth}{!}{%
\begin{tabular}{cccccccccc}
\hline    & \multicolumn{3}{c}{ \textbf{RSNA Pneumonia} } & \multicolumn{3}{c}{ \textbf{SIIM-ACR} } & \multicolumn{3}{c}{ \textbf{CheXpert} } \\
 \textbf{Methods} & $1 \%$ & $10 \%$ & $100 \%$ & $1 \%$ & $10 \%$ & $100 \%$ & $1 \%$ & $10 \%$ & $100 \%$ \\
\hline MoCo~\cite{moco} & $82.33$ & $85.22$ & $87.90$ & $75.49$ & $81.01$ & $88.43$ & $78.00$ & $86.27$ & $87.24$ \\
SimCLR~\cite{chen2020simple} & $80.18$ & $84.60$ & $88.07$ & $74.97$ & $83.21$ & $88.72$ & $67.41$ & $86.74$ & $87.97$ \\
\hline
ConVIRT~\cite{convirt} & $83.98$ & $85.62$ & $87.61$ & $84.17$ & $85.66$ & $91.50$ & $85.02$ & $87.58$ & $88.21$ \\
GLoRIA~\cite{gloria} & $84.12$ & $86.83$ & $89.13$ & $85.05$ & $88.51$ & $92.11$ & $83.61$ & $87.40$ & $88.34$ \\
BioViL~\cite{biovil} & $81.95$ & $85.37$ & $88.62$ & $79.89$ & $81.62$ & $90.48$ & $80.77$ & $87.56$ & $88.41$ \\
LoVT~\cite{lovt} & $85.51$ & $86.53$ & $89.27$ & $85.47$ & $88.50$ & $92.16$ & $85.13$ & $88.05$ & $88.27$ \\
PRIOR~\cite{prior} & $85.74$ & $87.08$ & $89.22$ & $87.27$ & $89.13$ & $92.39$ & $86.16$ & $88.31$ & $88.61$ \\
MedKLIP~\cite{medklip} & $87.31$ & $87.99$ & $89.31$ & $85.27$ & $90.71$ & $91.88$ & $86.24$ & $88.14$ & $88.68$ \\
\textbf{M\&M (Ours)} & \textbf{88.11} & \textbf{89.44} & \textbf{91.91} & \textbf{88.81} & \textbf{91.15} & \textbf{93.88} & \textbf{88.45} & \textbf{90.02} & \textbf{90.88} \\
\hline
\end{tabular}}
\caption{Comparing supervised classification results obtained by fine-tuning on RSNA Pneumonia Detection~\cite{shih2019augmenting}, SIIM~\cite{pneumo} and CheXpert~\cite{chexpert}. All methods are trained on different portions of the
training set from 1\% to 100\% and evaluated using AUC-ROC. Following PRIOR~\cite{prior} each reported value is the average of five runs along with the standard deviation.}
    \label{tab:supervised_classification}
\end{table*}

\begin{table*}[ht]
\centering
\resizebox{0.75\textwidth}{!}{%
\begin{tabular}{lccccccccc}
\hline
& \multicolumn{3}{c}{\textbf{RSNA Pneumonia}} & \multicolumn{3}{c}{\textbf{SIIM-ACR}} & \multicolumn{3}{c}{\textbf{NIH Chest X-Ray}} \\
\textbf{Methods} & AUC↑ & F1↑ & ACC↑ & AUC↑ & F1↑ & ACC↑ & AUC↑ & F1↑ & ACC↑ \\
\hline
ConVIRT~\cite{convirt} & 80.42 & 58.42 & 76.11 & 64.31 & 43.29 & 57.00 & 61.01 & 16.28 & 71.02 \\
GLoRIA~\cite{gloria} & 71.45 & 49.01 & 71.29 & 53.42 & 38.23 & 40.47 & 66.10 & 17.32 & 77.00 \\
BioViL~\cite{biovil} & 82.80 & 58.33 & 76.69 & 70.79 & 48.55 & 69.09 & 69.12 & 19.31 & 79.16 \\
PRIOR~\cite{prior} & 85.58 & 62.91 & 77.85 & 86.62 & 70.11 & 84.44 & 74.51 & 23.29 & 84.41 \\
MedKLIP~\cite{medklip} & 86.94 & 63.42 & 80.02 & 89.24 & 68.33 & 84.28 & 76.76 & 25.25 & 86.19 \\
\textbf{M\&M (Ours)} & \textbf{88.91} &  \textbf{66.58} & \textbf{83.14} & \textbf{91.15} & \textbf{71.58} & \textbf{86.15} & \textbf{77.92} & \textbf{27.55} & \textbf{88.52} \\
\hline
\end{tabular}}
\caption{ Comparing recent state-of-the-art methods on zero-shot classification task. We use AUC, F1 and ACC scores for comparison. Following MedKLIP~\cite{medklip} for evaluation on NIH Chest X-Ray, the metrics all refer to the macro average on the 14 diseases.}
\label{table:zero_class}
\end{table*}

\subsubsection{Zero-Shot}

We assess zero-shot classification on RSNA Pneumonia, SIIM-ACR, and NIH Chest X-Ray datasets with state-of-the-art models, showcasing our model's generalization across various clinical sources. Our approach achieves consistent performance enhancements, attributed to domain adaptation as discussed in MedKLIP~\cite{medklip}. Results in Table~\ref{table:zero_class} show our M\&M model outperforming existing methods by up to 3.16\%, following pre-training on MIMIC-CXR and up to a 3.29\% improvement on the novel COVID-19 disease in Table~\ref{tab:covid_description_comparison}.

\begin{table}[h!]
\centering
\begin{minipage}[t]{0.48\textwidth}
\centering
\begin{tabular}{cccc}
\hline
\textbf{Methods} & \textbf{AUC↑} & \textbf{F1↑} & \textbf{ACC↑} \\
\hline
ConVIRT~\cite{convirt} & 52.08 & 69.02 & 52.66 \\
GloRIA~\cite{gloria}  & 66.59 & 70.07 & 60.83 \\
BioViL~\cite{biovil}   & 53.82 & 69.10 & 53.75 \\
MedKLIP~\cite{medklip} & 73.96 & 76.70 & 70.06 \\
\textbf{M\&M (Ours)} & \textbf{75.15} & \textbf{77.89} & \textbf{73.35} \\
\hline
\end{tabular}
\caption{Performance on Covid-19 CXR.}
\label{tab:covid_description_comparison}
\end{minipage}\hfill
\begin{minipage}[t]{0.48\textwidth}
\centering
\begin{tabular}{lccc}
\hline
\textbf{Methods} & \textbf{AUC↑} & \textbf{F1↑} & \textbf{ACC↑} \\
\hline
ConVIRT~\cite{convirt}   & 77.00 & 56.76 & 69.19   \\
GLoRIA~\cite{gloria}    & 77.74 & 57.98 & 71.45  \\
BioViL~\cite{biovil}     & 75.40 & 55.72 & 69.14   \\
MedKLIP~\cite{medklip}     & 78.98 & 58.26 & 72.80   \\
\textbf{M\&M (Ours)} & \textbf{80.71} & \textbf{60.18} & \textbf{73.91}\\
\hline
\end{tabular}
\caption{Performance comparison on Edema severity grading multi-class task.}
\label{tab:grade}
\end{minipage}
\end{table}

\subsection{Grading}

In addition to diagnosis, determining disease severity is essential. We refine pre-trained features for a multi-category classification task using the Edema Severity dataset, which includes classes 0 to 3, each indicating a different severity stage. The average scores across all severity levels are shown in Table~\ref{tab:grade}. 

\subsection{Segmentation}

Table~\ref{tab:seg} presents our fine-tuning experiments for segmenting three distinct diseases, where we utilize 1\%, 10\%, and 100\% of the available data. Regardless of the varying image distributions associated with each disease, our techniques consistently outperform current leading methods. We see significant gains in particular when data is scarce outperforming previous works by up to 2.96\%.

\begin{table*}
\centering

\begin{tabular}{|l|ccc|ccc|ccc|}
\hline
\multirow{2}{*}{\textbf{Methods}} & \multicolumn{3}{c|}{\textbf{RSNA Pneumonia}} & \multicolumn{3}{c|}{\textbf{SIIM-ACR}} & \multicolumn{3}{c|}{\textbf{Covid-19}} \\ \cline{2-10} 
                                   & 1\%      & 10\%     & 100\%    & 1\%        & 10\%       & 100\%      & 1\%      & 10\%     & 100\%    \\ \hline
Scratch                            & 43.47   & 60.47   & 70.68   & 21.33     & 33.23     & 74.47     & 14.81   & 23.67   & 32.28   \\
ConVIRT~\cite{convirt}                           & 57.06   & 64.91   & 72.01   & 54.06     & 61.21     & 73.52     & 19.95   & 27.24   & 37.37   \\
GLoRIA~\cite{gloria}                            & 65.55   & 69.07   & 73.28   & 56.73     & 57.78     & 76.94     & 18.89   & 28.09   & 38.69   \\
BioVil~\cite{biovil}                             & 68.24   & 70.38   & 72.49   & 62.67     & 69.98     & 78.49     & 21.13   & 32.39   & 41.62   \\
PRIOR~\cite{prior}                               & 70.11   & 70.88   & 74.43   & 66.14     & 71.24     & 78.85   & 23.66  & 34.72   & 43.01   \\ 
MedKLIP~\cite{medklip}                               & 70.64   & 71.62   & 75.79   & 66.59     & 72.10     & 79.37    & 24.45   & 35.39   & 43.99   \\
\textbf{M\&M (Ours)} & \textbf{72.28} & \textbf{73.11} & \textbf{76.68} & \textbf{69.55} & \textbf{73.47} & \textbf{80.28} & \textbf{28.25} & \textbf{37.32} & \textbf{45.04} \\
\hline
\end{tabular}
\caption{Evaluating Dice scores against leading methods for segmentation tasks, we analyze diseases with 1\%, 10\%, 100\% labeled data, improving across all scenarios.}
\label{tab:seg}
\end{table*}

\subsection{Ablation Study}

We assess the effects of different image masking techniques—no masking, random patch masking~\cite{masked}, attention-guided masking (AttMask)\cite{attmask}, and salient patch selection (AutoMAE)\cite{automae}—and validate our manuscript generation approach. Our method boosts performance by as much as 8.06\%, as Table~\ref{tab:masking_strategy} illustrates. For report generation, we examine four methods: original reports, extracted triplets, knowledge-enhanced triplets (KE-Triplet) via MedKLIP, and converting triplets into manuscripts, using the same masking for all (Table~\ref{tab:manuscript}). KE-Triplet leads, surpassing even MedKLIP by up to 1.97\%, highlighting our model's efficacy on the NIH Chest X-Ray dataset.

\begin{table}[h!]
    \centering
    \begin{minipage}{0.48\textwidth}
        \centering
        \begin{tabular}{cccc}
            \hline
            \textbf{Methods} & \textbf{AUC↑} & \textbf{F1↑} & \textbf{ACC↑} \\
            \hline
            MaskVLM~\cite{maskvlm} & 58.87 & 14.96 & 66.69 \\
            \hline
            No Masking & 61.48 & 16.33 & 70.54 \\
            MAE~\cite{masked}  & 68.84 & 18.85 & 75.59 \\
            AttMask~\cite{attmask} & 71.85 & 20.84 & 77.81 \\
            AutoMAE~\cite{automae}   & 73.18 & 22.81 & 80.46 \\
            \textbf{M\&M (Ours)} & \textbf{77.92} & \textbf{27.55} & \textbf{88.52} \\
            \hline
        \end{tabular}
        \caption{Ablation study on masking strategy, focusing on zero-shot setting to directly assess learned features.}
        \label{tab:masking_strategy}
    \end{minipage}\hfill
    \begin{minipage}{0.48\textwidth}
        \centering
        \begin{tabular}{cccc}
            \hline
            \textbf{Methods} & \textbf{AUC↑} & \textbf{F1↑} & \textbf{ACC↑} \\
            \hline
            Original Report & 69.95 & 20.04 & 77.71 \\
            Triplet & 73.48 & 24.42 & 82.89 \\
            KE-Triplet & 76.84 & 26.11 & 86.55 \\
            \textbf{M\&M (Ours)} & \textbf{77.92} & \textbf{27.55} & \textbf{88.52} \\
            \hline
        \end{tabular}
        \caption{Ablation study on manuscript generation method selection, focusing solely on the zero-shot setting to evaluate the effectiveness of learned features.}
        \label{tab:manuscript}
    \end{minipage}
\end{table}

\section{Conclusion}

Medical contrastive learning is challenged by inconsistent semantics and morphology in sample text pairs, causing semantic drifts. Report variability from multiple authors further complicates semantic interpretation. To address this, we propose a two-step approach. First, standardize reports into triplets. Then, convert triplets into binary ``observations'' that guide towards ``verdicts''. For images, we use Meijering-based masking for pre-training instead of random masking to capture local context critical for medical images. Our multimodal contrastive learning framework, combining the standardized text and tailored visual representations, advances medical image analysis, achieving new state-of-the-art performances on multiple different downstream tasks. While the filter improves the performance in the discussed X-ray data, future work would be to test this on other modalities, such as MRI, which is noisier and more heterogeneous. 

\begin{credits}

\subsubsection{\discintname}
David A. Clifton was supported by the Pandemic Sciences Institute at the University of Oxford; the National Institute for Health Research (NIHR) Oxford Biomedical Research Centre (BRC); an NIHR Research Professorship; a Royal Academy of Engineering Research Chair; the Wellcome Trust funded VITAL project (grant 204904/Z/16/Z); the EPSRC (grant EP/W031744/1); and the InnoHK Hong Kong Centre for Cerebro-cardiovascular Engineering (COCHE).
\end{credits}

\bibliographystyle{splncs04}
\bibliography{main}

\end{document}